%% file: main.tex
\documentclass[12pt, a4paper]{article}
\usepackage{lrec2006}

\usepackage[fleqn]{amsmath}
\usepackage{amssymb}
\usepackage[utf8x]{inputenc}
\usepackage{t1enc}
\usepackage{graphicx}
\usepackage{tabularx}
\usepackage[T1]{url} 
\usepackage{booktabs}

\newcommand{\argmax}{\operatornamewithlimits{argmax}}

\usepackage{xspace}
\def\debut{\texttt{left}\xspace}
\def\fin{\texttt{right}\xspace}
\def\debutfin{\texttt{both}\xspace}
\def\dedans{\texttt{nothing}\xspace}

\begin{document}
\title{Learning Recursive Segments for Discourse Parsing}

\name{Stergos Afantenos$^{\ast}$, Pascal Denis$^{\dagger}$, Philippe Muller$^{\ast}\ ^{\dagger}$, Laurence Danlos$^{\dagger}$}

\address{$^{\ast}$Institut de recherche en informatique de Toulouse
  (IRIT)\\ Université Paul Sabatier \\ 118 route de Narbonne, 31062 Toulouse Cedex 9, France \\
  {\tt \{stergos.afantenos, philippe.muller\}@irit.fr} \\ \\
  $^{\dagger}$Equipe-Projet Alpage\\ INRIA \& Université
  Paris 7 \\ 30 rue Château des Rentiers, 75013 Paris, France \\
  {\tt laurence.danlos@linguist.jussieu.fr, pascal.denis@inria.fr}}

\abstract{Automatically detecting discourse segments is an important
  preliminary step towards full discourse parsing. Previous research
  on discourse segmentation have relied on the assumption that
  elementary discourse units (EDUs) in a document always form a linear
  sequence (i.e., they can never be nested). Unfortunately, this
  assumption turns out to be too strong, for some theories of
  discourse like SDRT allows for nested discourse units. In this
  paper, we present a simple approach to discourse segmentation that
  is able to produce nested EDUs. Our approach builds on standard
  multi-class classification techniques combined with a simple
  repairing heuristic that enforces global coherence. Our system was
  developed and evaluated on the first round of annotations provided
  by the French Annodis project (an ongoing effort to create a
  discourse bank for French). Cross-validated on only $47$ documents
  ($1,445$ EDUs), our system achieves encouraging performance results
  with an F-score of $73$\% for finding EDUs. }

\maketitleabstract

\section{Introduction}
\label{sec:intro}
% motivations
Discourse parsing is the analysis of a text from a global, structural
perspective: how parts of a discourse contribute to its global
interpretation, accounting for semantic and pragmatic effects beyond
simple sentence concatenation. This task consists in two main steps:
(i) finding the elementary discourse units (henceforth EDUs), and
(ii) organizing them in a way that make explicit their functional (aka
rhetorical) relations. Popular theories of discourse include
Rhetorical Structure Theory (RST) \cite{mann&thompson87}, Discourse
Lexicalized Tree-Adjoining Grammar (DLTAG)
\cite{DBLP:journals/cogsci/Webber04}, Segmented Discourse
Representation Theory (SDRT) \cite{asher93}. Each of these theoretical
frameworks has been at the center of important corpus building
efforts, see \cite{rst-corpus:03,pennDTB04,discor07} respectively. In
the present work, we focus on the first step, namely segmenting a
discourse into EDUs, within a larger project aiming at building an
SDRT discourse corpus of French texts.

In addition to being a necessary step in discourse parsing, discourse
segmentation, could also be useful as a stand-alone application for a
variety of other tasks where EDUs could provide simpler input than
sentences. Examples of such tasks are: automatic summarization and
sentence compression, bitext alignment, translation,
chunking/syntactic parsing.

% related work
The first discourse segmentation system dates back to the rule-based
work of \cite{ejerhed96}, which was a component in the RST-based
parser of \cite{marcu00}. More recently,
\cite{tofiloski-brooke-taboada:2009:Short} tested a rule-based
segmenter on top of a syntactic parser, achieving F-score of
$80$-$85$\% in segment boundary identification on a slightly modified
RST corpus. Machine learning based segmentation systems have also been
proposed, notably by \cite{soricut:marcu:03},
\cite{sporleder:lapata:05} and \cite{fisher:roark:07}. The latter
report F-score of $90.5$\% in boundary detection (and $85.3$\% in
correct bracketing) on the RST corpus.

Discourse segmentation is to large extent theory dependent, for
different theories make different assumptions on what EDUs can
be. Carried out on the RST corpus, previous work on discourse
segmentation has exploited an important particularity of this corpus:
namely, the fact that it does not have any embedded EDUs. These
approaches have been able to recast discourse segmentation as a
\textit{binary} classification problem: that is, each text position
(token or token separator) is either a segment boundary or not. By
contrast to RST, other theories like SDRT allows for embedded EDUs:
embedding is used to encode modifying clauses like non restrictive
relatives (including reduced relatives) and appositions. As will be
discussed in Section~\ref{sec:corpus}, our SDRT-based corpus does
contain close to $10$\% of nested EDUs.

Predicting nested structures introduces additional difficulties, in
particular that of outputting a coherent, balanced bracketing. This
characteristic renders discourse segmentation akin to syntactic clause
boundary identification (CBI), a task which has received some
attention from the CL community. The main approach to CBI is to
classify tokens into \textit{three} classes for clause start, end, or
inside. The best results obtained during the CoNLL-2001 campaign were
$89$-$90$\% for boundary detection and $81.73$\% for correct clause
identification (correct guessing of start and end), with boosted
decision trees \cite{carreras2001conll}.

% + recent: sur le coréen (pricai 2006), avec CRF (cicling
% 2008)[+regles linguistiques], résultats similaires (80-89)

%%% our contribution
We have adapted this general setting to the problem of discourse
segmentation, with possible embedded segments, and applied it to a
corpus of French discourses, part of an on-going corpus building
project.%, Annodis.

% NOTES on this section: 1. What is the level of familiarity that the
% annotators have had with SDRT?  2. Given the page constraints of
% LREC, do we mention anything at all concerning the manual that we
% have provided them? If yes, to which level of detail?  3. human
% annotation agreement??

\section{Data and Evaluation}
\label{sec:corpus}

\subsection{Corpus}
The corpus we use has been developed as part of the Annodis
project\footnote{\url{http://w3.erss.univ-tlse2.fr/textes/pagespersos/annodis/}},
an on-going effort to annotate French discourses from various genres
with both top-level, typographic structures and local coherence
relations. About $100$-$150$ texts are being segmented and annotated
with coherence relations. These documents are drawn mainly for
wikipedia articles and from \textit{L'est républicain} newspaper
\footnote{\url{http://www.cnrtl.fr/corpus/estrepublicain/}}. Text
length varies from $300$ to $900$ tokens. Annotations are performed by
pairs of human annotators in a two-step process: (i) individual
annotations, and (ii) adjudication. The present work considers the
$47$ texts that have undergone validation. The average number of EDUs
per document in this set is $33$.

%\pacomment{On a vraiment besoin d'une ref pour Annodis! p-e le papier
%  TALN de l'an dernier??}
%\phcomment{nan poster en français ... il nous faudrait un TR
%sur le manuel d'annotation}

%\pacomment{Lien avec la SDRT: jeu de relations, etc. + annotator
%  agreement scores?}

% XX description qualitative des segments: synthèse du manuel XXX
% sinon l'article est vraiment trop abstrait
Segments typically correspond to verbal clauses, but also other
syntactic units describing eventualities (such as prepositional
phrases), adjuncts such as appositions or cleft constructions with
discursive long-range effects such as frame adverbials.
%The specification of the task is detailed in a set of guidelines
%aiming at annotator training.
%\pacomment{Ref pour le guide d'annotation?}
A particularity of the discourse units in Annodis is that they can be
embedded in one another, as in example in figure~1 (brackets mark segmentation).

\begin{figure*}
  \label{fig:nested_exple}
  {\sf [Ces pièces, [mondialement connues,]$_{\pi_1}$ [donc
    difficilement écoulables,]$_{\pi_2}$ avaient été repérées chez un
    riche amateur nippon]$_{\pi_0}$}\\[+.5cm]
  {\it [The pieces, [worldwide famous,]$_{\pi_1}$ [thus hard to
    resell,]$_{\pi_2}$ had been located at a rich japanese art
    lover's]$_{\pi_0}$}
  \caption{A discourse segmentation from the Annodis corpus.}
\end{figure*}

In this example, the EDUs $\pi_1$ \textit{mondialement connues,} and
$\pi_2$ \textit{donc difficilement écoulables,} are nested within the
the main, discontinuous EDU $\pi_0$ \textit{Ces pièces avaient été
  repérées chez un riche amateur nippon}.

%\pacomment{Etre plus explicite sur le probleme des EDUs discontinues
%  et sur la justification du nesting?}

% written in the French language and consists of 47 documents, 15 of
% which have been taken from the Wikipedia while the rest have

%\begin{table}%\centering
%  \begin{tabular}{ccc}
%    \toprule
%    Documents & EDUs per Document & Words per Document\\
%    \midrule
%    47 & about 33 & about 272\\
%    \bottomrule
% \end{tabular}
%\caption{Several statistics of the corpus \label{tab:corpusStats}}
%\end{table}

\subsection{Evaluation}

Discourse segmentation evaluation is typically performed in terms of
precision, recall, and F-score for segment boundaries
\cite{soricut:marcu:03,fisher:roark:07,sporleder:lapata:05}. Previous
work differ as to whether they include sentence boundaries (e.g.,
\cite{soricut:marcu:03} are only interested in sentence-internal
segmentation) and whether they additionally require labeling of the
segments \cite{sporleder:lapata:05}.

Since the type of segmentation we produce includes nested EDUs, we
have to resort to another type of evaluation. For this paper, we use
the three metrics commonly used for evaluating clause detection: (i)
precision, recall, and F-score for segment start position, (ii)
precision, recall, and F-score for segment end position, and (iii)
precision, recall, and F-score for complete segments. These metrics
correspond to three tasks included in the CoNLL 2001 shared task.

\section{Approach}

%\pacomment{Lien avec l'exemple...}

\subsection{Classification Model}
Like previous approaches to discourse segmentation and CBI, we cast
the task of EDU identification as a classification
problem. Specifically, we built a \textit{four}-class classifier that
maps each token $w_i$ in a discourse $w_1,\ldots,w_n$ to one of the
following boundary types $B=\{\debut,\fin,\debutfin,\dedans\}$. These
correspond to the different bracketing configurations found in our
corpus, respectively (i) $w_i$ opens a segment, (ii) $w_i$ ends a
segment, (iii) $w_i$ is a single-token segment, and (iv) none of the
above. If we take the beginning of the example in
\ref{fig:nested_exple}, {\it [Ces pièces, [mondialement connues,] }
{\it Ces} and {\it mondialement} would be classified as $\debut$, the
last comma as $\fin$, and all other tokens as $\dedans$.

For our classifier, we used a regularized maximum entropy (MaxEnt, for
short) model \cite{berger_et_al:96}. In MaxEnt, the parameters of an
exponential model of the following form are estimated:
\begin{equation*}
  P(b|t) = \frac{1}{Z(b)} \exp\left(\sum\limits_{i=1}^m w_i f_i(t,b)\right)
\end{equation*}
where $t$ represents the current token and $b$ the outcome (i.e.,
the type of boundary). Each token $t$ is encoded as a vector of $m$
indicator features $f_i$. There is one weight/parameter $w_i$
for each feature $f_i$ that predicts its classification
behavior. Finally, $Z(b)$ is a normalization factor over the different
class labels (in this case, the $4$ boundary types), which guarantees
that the model outputs probabilities.

In MaxEnt, the values for the different parameters $\hat{w}$ are
obtained by maximizing the log-likelihood of the training data $T$
with respect to the model \cite{berger_et_al:96}:
\begin{equation*}
  \hat{w} = \argmax_{w} \sum_{i}^{T} \log P(b^{(i)}|t^{(i)})
\end{equation*}
Various algorithms have been proposed for performing parameter
estimation (see \cite{malouf:02} for a comparison). Here, we used the
Limited Memory Variable Metric Algorithm implemented in the MegaM
package.\footnote{Available from
  \url{http://www.cs.utah.edu/~hal/megam/}.} We used the default
regularization prior that is used in MegaM.

\subsection{Feature Set}
%\label{sec:features}

Our feature set relies on two main sources of information. The first
source is a list of \textit{lexical markers}, containing discourse
connectives and a few indirect speech report verbs that are likely to
introduce discourse units. Specifically, we created boolean features
that check whether the token is part of connectives (resp. verbs) in
our list of markers.

The other information source is (morpho-)syntactic, drawn from the
automatic analysis provided by the Macaon chunker \cite{nasr:2006b}
and the \textsc{Syntex} dependency parser \cite{syntex05}. Using these
two analyzers, we extract for each token: its lemma, its
part-of-speech (POS) tag, its chunk tag, its dependency path to the
root element (as well as ``sub-paths'' of length $1$-$3$), and its
inbound dependencies. In addition, we also capture the linear position
of the word in a sentence (we used quantized values ranging from
$1$-$100$). These feature templates were also applied to the
surrounding words in a window of $3$ words to the left and right.

Two more feature families were added. The first concerns the outward
chunk sequence for each token; that is, given that a token is embedded
in a sequence of chunks, we start from the innermost chunk tag and we
go out all the way to the outermost. These features exploit the fact
that Macaon provides some level of embedding in its chunks.  The
second feature family concerns all the n-gramms $1<n\leq6$ for which
the token is included and their span does not exceed the boundaries of
the current sentence. A synoptic table with the entire feature set we
used is shown in table~\ref{table:features2}.

\begin{table*}\centering
  \begin{tabular}{l|l}
    \toprule
    Feature & Description  \\
    \midrule
    Lemma    & the token's lemma (Syntex) \\
    POS & Part of speech (Macaon)\\
    Grammatical category & the main grammatical category of the token: V, N, P, etc. (Syntex)\\
    start of a discourse marker & boolean, indicating whether the tokens starts a discourse marker\\
    indirect speech report verb & boolean, indicating whether the token belongs to a predefined\\
        &  list of verbs.\\
    dependency path & the dependency path from the word towards the root, limited\\
        &  to distance 3 (Syntex)\\
    inbound dependencies  & the inbound dependency relations for each token (Syntex)\\
    syntactic projections & the number of times that the token is at the start, end or middle\\
          &  of an NP, VP, PP projection (Syntex)\\
    distance from sentence boundaries & the relative distance from each of the sentence boundaries\\
    context 3-grams & the lemma and POS 3-grams before and after the \\
        & token (Syntex \& Macaon)\\
    chunk start/end & boolean features; token coincides with a chunk start/end (Macaon)\\
    outward chunk tag sequence & the sequence of chunk tags from the innermost to the \\
        & outermost chunk (Macaon)\\
    context n-gramms & all the n-gramms ($1<n\leq6$) that include the token and do\\
        &  not exceed the  limits of the sentence. The n-grams include\\
         & Lemmas (Syntex), POS tags (Macaon) and Chunk tags (Macaon)\\
    \bottomrule
 \end{tabular}
\caption{Features used for the second approach (including chunks).} \label{table:features2}
\end{table*}

\subsection{Resampling}

The distribution of boundary types is heavily skewed towards \dedans (about 12.000 instances against about 1400 for each \debut and \fin),
which suggests that resampling our data toward a more uniform
distribution might lead to better classification accuracy, and in turn
to better EDU segmentation.

%\pacomment{Add the class distribution if available!}

The resampling method we used directly exploits the syntactic chunk
boundaries as found by the Macaon chunker. It is based on the
observation that EDU boundaries in a large majority of cases coincide
with chunk boundaries. The output of Macaon was used in the following
ways. First, we decided to replace the decisions on sentence boundary
tokens with the decisions that Macaon provides. In other words,
sentence boundary tokens, as given by Macaon, were ignored during
training; they were tagged as \debut and \fin respectively during
test. Second, we also removed from training tokens that were
\textit{strictly} inside chunks (that is, tokens that are inside a
chunk but doesn't correspond to its beginning or end). At test, these
tokens were assigned the \dedans class. All remaining tokens were used
for training and follow the classification decoding at test. After those modifications, the class distribution was around 9200 instances for the class \dedans, while the rest of the classes had around 1400 instances.

% This decision has had the side effect that all instances of
% the \debutfin class have been eliminated from our pool, since those
% instances concerned single-token EDUs found in the articles' section
% headings. By consequence, we limited our classes into the following
% three $C=\{\debut,\fin,\dedans\}$ completely eliminating the highly
% skewed class \debutfin.
%\pacomment{Ideally, we would like to include the class distrib after
%  resampling}

\subsection{Enforcing coherence}
% Given that we aim at predicting nested structures, we also tackle the
% problem of ensuring that our local classification decisions ultimately
% produce a coherent structure. This section describes our
% classification scheme, the feature set used, and a simple technique
% that attempts to correct local classifications yielding incoherent
% bracketings.

Casting segmentation as a series of local classifications has two
major drawbacks. First, the segmentation decision at a token is highly
dependent from the decisions on neighboring tokens. Secondly,
unrelated local decisions do not guarantee the well-formedness of the
segmentation of a sentence, since we allow for embedded segments. For
instance, the number of beginning of embedded segments must obviously
match the number of endings.

A straightforward way to capture Markovian dependencies between
segmentation labels is to encode previous labels as features of the
model, in combination with a Viterbi decoding. Unfortunately, we found
during development that this strategy degrades segmentation
performance, probably due to the sparsity of the boundary
labels.\footnote{Similar findings are reported by
  \cite{fisher:roark:07}.}

To tackle the problem of ensuring a coherent bracketing, we propose a
specific post-processing on the outputs of the classifier. In
particular, we apply heuristic repair techniques (adding/deleting
boundaries) to yield a well-formed sentence segmentation.  A simple
technique proved efficient enough: we scanned sentences token by token
from beginning to end, while keeping track of the depth of the current
EDU embedding.  If the depth is 0 before the end of a sentence, it
means we found a stranded token, that is then reclassified as
$\debut$; this rebalances the number of $\debut$ and $\fin$.  Dually,
we reversed the sequences to reclassify remaining out-of-segment
tokens as $\fin$.  This heuristic is illustrated in
figure 2. In the future we plan to apply local
optimization techniques under well-formedness constraints, to repair
segmentations while better preserving the probability on each
decision.

\begin{figure*}
  \label{fig:post-proc}
 {\bf Input from classifier}: \\{\it [The pieces,] worldwide famous,] thus hard to
    resell,] had been located [at a rich japanese art
    lover's]}\\[+.25cm]
{\bf First pass left-to-right}: \\ {\it [The pieces,] \underline{[}worldwide famous,] \underline{[}thus hard to
    resell,] \underline{[}had been located [at a rich japanese art
    lover's]}\\[+.25cm]
{\bf First pass right-to-left}:  {\it [The pieces,] [worldwide famous,] [thus hard to
    resell,] [had been located\underline{]} [at a rich japanese art
    lover's]}\\[+.25cm]
  \caption{Example repairing of a not well-formed segmentation with additions underlined. The sentence can now be compared to the reference, cf figure~1.}
\end{figure*}

%\pacomment{Si possible, comparer avec l'heuristique de Carreras et
%  Marques.}

\section{Experiments and Results}\label{sec:expes}

We present two sets of scores, one without post-processing and one
with post-processing. We did a $10$-fold cross-validation on the
sentences contained in the $47$ documents of the corpus. We used the
three metrics for segmentation evaluation discussed in section
\ref{sec:corpus}; we also report precision, recall, and F-score for
the \debutfin boundary class.

Table~\ref{table:eval1} (resp. table~\ref{table:eval2}) reports the
performance scores of the ``classifier-only'' system
(resp. ``classifier+post-processing'' system) for the first series of
experiments. In terms of overall classification performance, both
systems perform similarly, but the second system improves on the three
boundary classes $\{\debut,\fin,\debutfin\}$. The main source of
improvement comes from recall, which suggests that our heuristics
recover boundaries that were missed by the classifier.

%\pacomment{!!!!!!!Update scores!!!!!!}
%\phcomment{done for repair}

Before post-processing, the proportion of not well-formed
segmentations on the (recognized) sentences is $35$\%, our
post-processing heuristics yield $98$\% well-formed segmentations.
The impact on precision/recall is shown in table \ref{table:eval2}.

The overall bad performance on $\debutfin$ is due to the lack of data
for this class: there are less than $20$ examples in the entire
corpus. When it comes to the segment evaluation, again the best results were
achieved by the second approach which managed to correctly identify
$73$\% of the manually annotated segments. These results are slightly
less, but close to, the best results obtained by systems on the CBI
task. Of course, the main reason post-processing boosts the EDU score
is that a third more of the sentences are now evaluated, since they
are well-formed. But the decline in precision is much less than the
gain in recall.

%  \pacomment{IMPORTANT: Mettre a jour la derniere phrase avec les
%    scores de Stergos. On aurait p-e du faire un baseline avec la
%    classe la + courante?}

\begin{table}\centering
  \begin{tabular}{l||c|c|c}
    \toprule
    Class & Recall & Precision & F-measure  \\
    \midrule
    Left    & 0.845 & 0.891 & 0.868  \\
    Right   & 0.881 & 0.925 & 0.902  \\
    Both    & 0.684 & 0.812 & 0.742 \\
    %Nothing & 0.971 & 0.953 & 0.962  \\
    \midrule
    EDUs   & 0.427 & 0.880 & 0.575  \\
    \bottomrule
 \end{tabular}
\caption{Evaluation without post-processing.} \label{table:eval1}
\end{table}

%         NOTHING |      0.971      0.953      0.962 |      10325
%   10836      10636
%                BOTH |      0.294      0.714      0.417 |          5
%       7         17
%               RIGHT |       0.78      0.855      0.816 |        939
%    1098       1204
%                LEFT |      0.793      0.851      0.821 |        979
%    1150       1234
% ------------------------------------------------------------------------------------------

% Overall Acc: 0.936

\begin{table}\centering
  \begin{tabular}{l||c|c|c}
    \toprule
    Class & Recall & Precision & F-measure  \\
    \midrule
    Left    & 0.876 & 0.880 & 0.878  \\
    Right   & 0.885 & 0.889 & 0.888  \\
    Both    & 0.684 & 1.0   & 0.812 \\
    % Nothing & 0.961 & 0.961 & 0.961  \\
    \midrule
    EDUs   & 0.719 & 0.748 & 0.733  \\
    % Total   & 0.935 & 0.935 & 0.935  \\
    \bottomrule
 \end{tabular}
 \caption{Evaluation with post-processing.} \label{table:eval2}
\end{table}

\subsection{Learning Curve}
For their RST EDU segmentation experiments, \newcite{fisher:roark:07}
have been using the RST-DT corpus which consists of a total of $385$
documents ($176,000$ tokens). \newcite{carreras2001conll} have used
the CoNLL 2001 corpus for the task of clause boundaries
identification: this corpus includes sections $15-18$ of the Penn
Treebank for training ($211,727$ tokens) and section $20$ for test
($47,377$ tokens). In contrast to those approaches we have worked, as
mentioned in section~\ref{sec:corpus} we have been working with $47$
validated documents ($14384$ tokens) from the Annodis project. Given
that the number of documents that we have been working with is
limited, at least in comparison with other approaches, we have
calculated the learning curve for this number of documents in order to
understand how the learning procedure will be influenced once we have
the totality of our documents annotated. As mentioned in
section~\ref{sec:corpus} the total number of documents expected will
be in the range of $100$ to $150$.

\begin{figure*}[t]
  \centering
  \input{learning_curve_full.latex}
  \caption{Learning curve for the series of experiments including chunking.}\label{fig:lc}
\end{figure*}
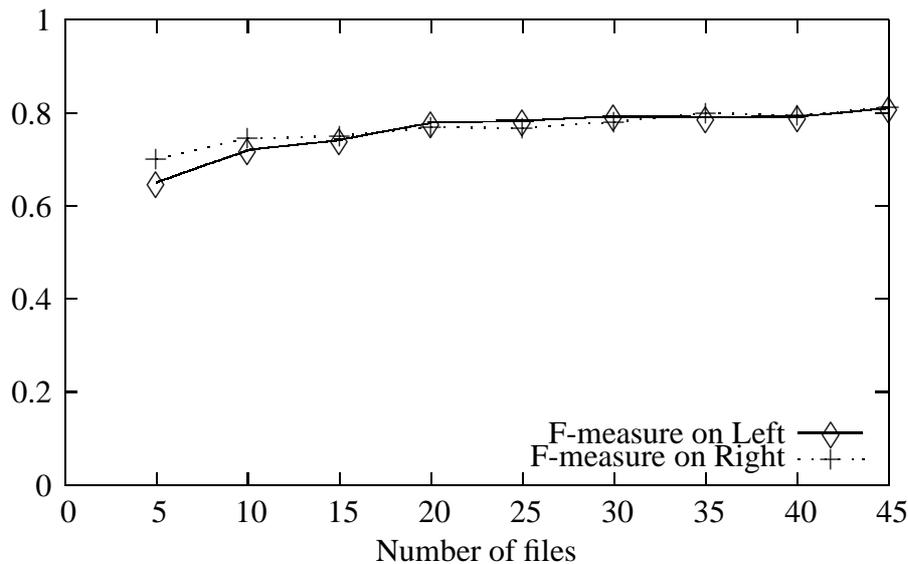

%\pacomment{Add number of tokens in our corpus for comparison sake}

In order to calculate our learning curve, we divided our corpus into
$9$ different learning sets, starting from $5$ random documents and
incrementally adding $5$ random documents into each learning set. For
each such set we performed a ten-fold cross-validation, in the same
way as described in section~\ref{sec:expes}, using the feature set
shown in table~\ref{table:features2}.

The learning curve is shown in figure~3. As it can be seen
from this figure, the curves for both classes (\debut and \fin)
grow regularly between sets $5$ to $30$ while it seems to plateau between
sets $30$ and $40$ only to start going up again during the last set of
documents. In general, it seems that the addition of more documents
will only slightly increase the performance of our approach.

\section{Conclusions and Future Work}
Discourse segmentation is a crucial preprocessing stage for discourse
analysis, and the global reliability of discourse parsing is heavily
determined by success at this level. We have proposed a simple
approach combining a 3-class classifier with a post-processing
heuristics that achieve reasonable results, although the data
available at the moment is limited. We need to see how this
generalizes to the whole corpus, and to check how dependent it is on
the nature of the corpus (newspaper articles and encyclopedia article).
% (PM) only in LREC ! :
Another angle we plan to investigate is the usefulness of a
non-perfect segmentation to help annotators start discourse
annotation. Given the cost of human annotation of discourse, saving
time on the segmentation would be a boost to annotators productivity,
provided we verify that time spent is roughly proportional to the
number of errors in the automated preprocessing; that hypothesis is
not necessarily true, and there might be a threshold on the precision
of the processing that is acceptable.  Mainly, the ideal trade-off
between precision and recall remains to be investigated.

\bibliographystyle{lrec2006}
\bibliography{bib_segmentation}
\end{document}

%% file: learning_curve_full.latex
% GNUPLOT: LaTeX picture
\setlength{\unitlength}{0.240900pt}
\ifx\plotpoint\undefined\newsavebox{\plotpoint}\fi
\sbox{\plotpoint}{\rule[-0.200pt]{0.400pt}{0.400pt}}%
\begin{picture}(1500,900)(0,0)
\sbox{\plotpoint}{\rule[-0.200pt]{0.400pt}{0.400pt}}%
\put(170.0,131.0){\rule[-0.200pt]{4.818pt}{0.400pt}}
\put(150,131){\makebox(0,0)[r]{ 0}}
\put(1430.0,131.0){\rule[-0.200pt]{4.818pt}{0.400pt}}
\put(170.0,277.0){\rule[-0.200pt]{4.818pt}{0.400pt}}
\put(150,277){\makebox(0,0)[r]{ 0.2}}
\put(1430.0,277.0){\rule[-0.200pt]{4.818pt}{0.400pt}}
\put(170.0,423.0){\rule[-0.200pt]{4.818pt}{0.400pt}}
\put(150,423){\makebox(0,0)[r]{ 0.4}}
\put(1430.0,423.0){\rule[-0.200pt]{4.818pt}{0.400pt}}
\put(170.0,568.0){\rule[-0.200pt]{4.818pt}{0.400pt}}
\put(150,568){\makebox(0,0)[r]{ 0.6}}
\put(1430.0,568.0){\rule[-0.200pt]{4.818pt}{0.400pt}}
\put(170.0,714.0){\rule[-0.200pt]{4.818pt}{0.400pt}}
\put(150,714){\makebox(0,0)[r]{ 0.8}}
\put(1430.0,714.0){\rule[-0.200pt]{4.818pt}{0.400pt}}
\put(170.0,860.0){\rule[-0.200pt]{4.818pt}{0.400pt}}
\put(150,860){\makebox(0,0)[r]{ 1}}
\put(1430.0,860.0){\rule[-0.200pt]{4.818pt}{0.400pt}}
\put(170.0,131.0){\rule[-0.200pt]{0.400pt}{4.818pt}}
\put(170,90){\makebox(0,0){ 0}}
\put(170.0,840.0){\rule[-0.200pt]{0.400pt}{4.818pt}}
\put(312.0,131.0){\rule[-0.200pt]{0.400pt}{4.818pt}}
\put(312,90){\makebox(0,0){ 5}}
\put(312.0,840.0){\rule[-0.200pt]{0.400pt}{4.818pt}}
\put(454.0,131.0){\rule[-0.200pt]{0.400pt}{4.818pt}}
\put(454,90){\makebox(0,0){ 10}}
\put(454.0,840.0){\rule[-0.200pt]{0.400pt}{4.818pt}}
\put(597.0,131.0){\rule[-0.200pt]{0.400pt}{4.818pt}}
\put(597,90){\makebox(0,0){ 15}}
\put(597.0,840.0){\rule[-0.200pt]{0.400pt}{4.818pt}}
\put(739.0,131.0){\rule[-0.200pt]{0.400pt}{4.818pt}}
\put(739,90){\makebox(0,0){ 20}}
\put(739.0,840.0){\rule[-0.200pt]{0.400pt}{4.818pt}}
\put(881.0,131.0){\rule[-0.200pt]{0.400pt}{4.818pt}}
\put(881,90){\makebox(0,0){ 25}}
\put(881.0,840.0){\rule[-0.200pt]{0.400pt}{4.818pt}}
\put(1023.0,131.0){\rule[-0.200pt]{0.400pt}{4.818pt}}
\put(1023,90){\makebox(0,0){ 30}}
\put(1023.0,840.0){\rule[-0.200pt]{0.400pt}{4.818pt}}
\put(1166.0,131.0){\rule[-0.200pt]{0.400pt}{4.818pt}}
\put(1166,90){\makebox(0,0){ 35}}
\put(1166.0,840.0){\rule[-0.200pt]{0.400pt}{4.818pt}}
\put(1308.0,131.0){\rule[-0.200pt]{0.400pt}{4.818pt}}
\put(1308,90){\makebox(0,0){ 40}}
\put(1308.0,840.0){\rule[-0.200pt]{0.400pt}{4.818pt}}
\put(1450.0,131.0){\rule[-0.200pt]{0.400pt}{4.818pt}}
\put(1450,90){\makebox(0,0){ 45}}
\put(1450.0,840.0){\rule[-0.200pt]{0.400pt}{4.818pt}}
\put(170.0,131.0){\rule[-0.200pt]{0.400pt}{175.616pt}}
\put(170.0,131.0){\rule[-0.200pt]{308.352pt}{0.400pt}}
\put(1450.0,131.0){\rule[-0.200pt]{0.400pt}{175.616pt}}
\put(170.0,860.0){\rule[-0.200pt]{308.352pt}{0.400pt}}
\put(810,29){\makebox(0,0){Number of files}}
\put(1290,213){\makebox(0,0)[r]{F-measure on Left}}
\put(1310.0,213.0){\rule[-0.200pt]{24.090pt}{0.400pt}}
\put(312,605){\usebox{\plotpoint}}
\multiput(312.00,605.58)(1.398,0.498){99}{\rule{1.214pt}{0.120pt}}
\multiput(312.00,604.17)(139.481,51.000){2}{\rule{0.607pt}{0.400pt}}
\multiput(454.00,656.58)(4.562,0.494){29}{\rule{3.675pt}{0.119pt}}
\multiput(454.00,655.17)(135.372,16.000){2}{\rule{1.838pt}{0.400pt}}
\multiput(597.00,672.58)(2.657,0.497){51}{\rule{2.204pt}{0.120pt}}
\multiput(597.00,671.17)(137.426,27.000){2}{\rule{1.102pt}{0.400pt}}
\multiput(739.00,699.61)(31.495,0.447){3}{\rule{19.033pt}{0.108pt}}
\multiput(739.00,698.17)(102.495,3.000){2}{\rule{9.517pt}{0.400pt}}
\multiput(881.00,702.59)(10.790,0.485){11}{\rule{8.214pt}{0.117pt}}
\multiput(881.00,701.17)(124.951,7.000){2}{\rule{4.107pt}{0.400pt}}
\put(1023,707.17){\rule{28.700pt}{0.400pt}}
\multiput(1023.00,708.17)(83.432,-2.000){2}{\rule{14.350pt}{0.400pt}}
\put(1166,706.67){\rule{34.208pt}{0.400pt}}
\multiput(1166.00,706.17)(71.000,1.000){2}{\rule{17.104pt}{0.400pt}}
\multiput(1308.00,708.58)(5.198,0.494){25}{\rule{4.157pt}{0.119pt}}
\multiput(1308.00,707.17)(133.372,14.000){2}{\rule{2.079pt}{0.400pt}}
\put(312,605){\raisebox{-.8pt}{\makebox(0,0){$\Diamond$}}}
\put(454,656){\raisebox{-.8pt}{\makebox(0,0){$\Diamond$}}}
\put(597,672){\raisebox{-.8pt}{\makebox(0,0){$\Diamond$}}}
\put(739,699){\raisebox{-.8pt}{\makebox(0,0){$\Diamond$}}}
\put(881,702){\raisebox{-.8pt}{\makebox(0,0){$\Diamond$}}}
\put(1023,709){\raisebox{-.8pt}{\makebox(0,0){$\Diamond$}}}
\put(1166,707){\raisebox{-.8pt}{\makebox(0,0){$\Diamond$}}}
\put(1308,708){\raisebox{-.8pt}{\makebox(0,0){$\Diamond$}}}
\put(1450,722){\raisebox{-.8pt}{\makebox(0,0){$\Diamond$}}}
\put(1360,213){\raisebox{-.8pt}{\makebox(0,0){$\Diamond$}}}
\put(1290,172){\makebox(0,0)[r]{F-measure on Right}}
\multiput(1310,172)(20.756,0.000){5}{\usebox{\plotpoint}}
\put(1410,172){\usebox{\plotpoint}}
\put(312,642){\usebox{\plotpoint}}
\multiput(312,642)(20.248,4.563){8}{\usebox{\plotpoint}}
\multiput(454,674)(20.747,0.580){6}{\usebox{\plotpoint}}
\multiput(597,678)(20.655,2.036){7}{\usebox{\plotpoint}}
\multiput(739,692)(20.753,-0.292){7}{\usebox{\plotpoint}}
\multiput(881,690)(20.704,1.458){7}{\usebox{\plotpoint}}
\multiput(1023,700)(20.657,2.022){7}{\usebox{\plotpoint}}
\multiput(1166,714)(20.747,-0.584){7}{\usebox{\plotpoint}}
\multiput(1308,710)(20.669,1.892){7}{\usebox{\plotpoint}}
\put(1450,723){\usebox{\plotpoint}}
\put(312,642){\makebox(0,0){$+$}}
\put(454,674){\makebox(0,0){$+$}}
\put(597,678){\makebox(0,0){$+$}}
\put(739,692){\makebox(0,0){$+$}}
\put(881,690){\makebox(0,0){$+$}}
\put(1023,700){\makebox(0,0){$+$}}
\put(1166,714){\makebox(0,0){$+$}}
\put(1308,710){\makebox(0,0){$+$}}
\put(1450,723){\makebox(0,0){$+$}}
\put(1360,172){\makebox(0,0){$+$}}
\put(170.0,131.0){\rule[-0.200pt]{0.400pt}{175.616pt}}
\put(170.0,131.0){\rule[-0.200pt]{308.352pt}{0.400pt}}
\put(1450.0,131.0){\rule[-0.200pt]{0.400pt}{175.616pt}}
\put(170.0,860.0){\rule[-0.200pt]{308.352pt}{0.400pt}}
\end{picture}